\documentclass{article}

\usepackage{arxiv}

\usepackage[utf8]{inputenc} 
\usepackage[T1]{fontenc}    
\usepackage{url}            
\usepackage{booktabs}       
\usepackage{nicefrac}       
\usepackage{microtype}      

\usepackage{amsmath,amsfonts,amssymb}
\usepackage{graphicx}
\usepackage{csquotes}
\usepackage{multirow}
\usepackage[colorlinks=true, allcolors=blue]{hyperref}
\usepackage{cleveref}

\title{Network transferability of adversarial patches in real-time object detection}

\author{
    Jens Bayer \\
    Fraunhofer Center for Machine Learning,\\
    Fraunhofer IOSB\\
    Gutleuthausstr. 1, Ettlingen, Germany\\
    jens.bayer@iosb.fraunhofer.de
\And
Stefan Becker\\
    Fraunhofer IOSB\\
    Gutleuthausstr. 1, Ettlingen, Germany\\
\And
David M\"unch\\
    Fraunhofer IOSB\\
    Gutleuthausstr. 1, Ettlingen, Germany\\
\And
Michael Arens\\
    Fraunhofer IOSB\\
    Gutleuthausstr. 1, Ettlingen, Germany\\
}

\begin{document} 
\maketitle

\begin{figure}[htb]
    \vspace{-1.5cm}
    \centering
    \includegraphics[width=0.6\textwidth]{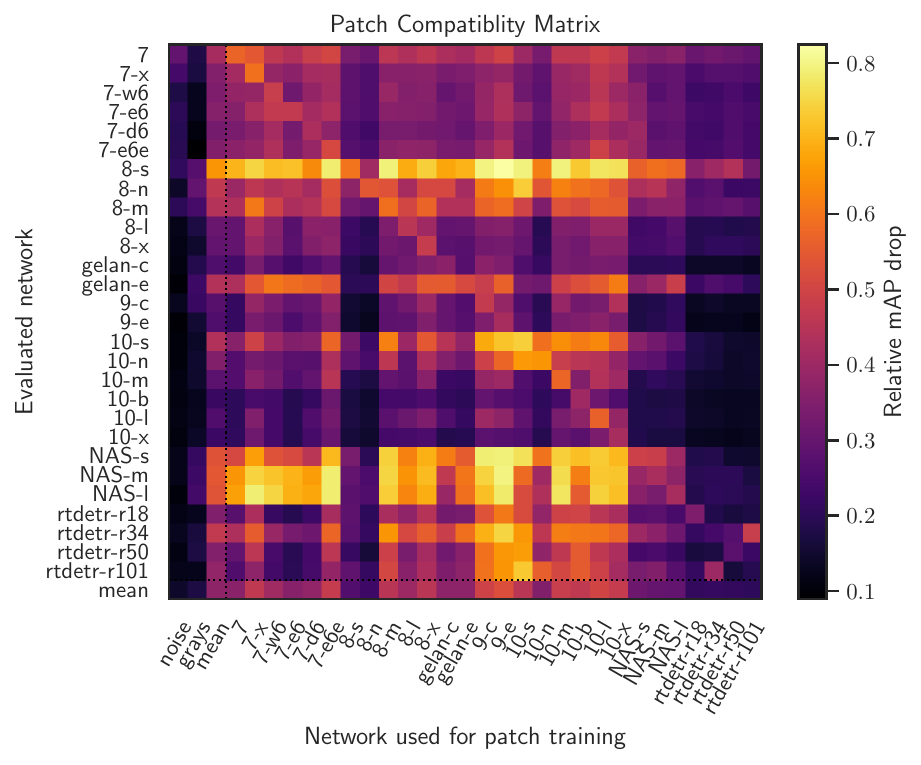}
    \caption{Compatibility matrix of the networks evaluated on the INRIA
    Person test set. Each cell represents the mean average precision (mAP) drop
    of a set of patches on the test set for a specific network. The brighter the
    color, the higher the relative mAP drop of the patch. }
    \label{fig:compatibility-matrix}
\end{figure}
\begin{abstract}
Adversarial patches in computer vision can be used, to fool deep neural networks
and manipulate their decision-making process. One of the most prominent examples
of adversarial patches are evasion attacks for object detectors. By covering
parts of objects of interest, these patches suppress the detections and thus
make the target object \enquote{invisible} to the object detector. Since these
patches are usually optimized on a specific network with a specific train
dataset, the transferability across multiple networks and datasets is not given.
This paper addresses these issues and investigates the transferability across
numerous object detector architectures. Our extensive evaluation across
various models on two distinct datasets indicates that patches optimized with
larger models provide better network transferability than patches that are
optimized with smaller models.
\end{abstract}


\section{Introduction and Related Work}
As deep neural networks become increasingly mature and get included in everyday
life, there is a growing demand for robust yet fast and accurate models.
Robustness refers to the ability of a model to correctly handle new,
corrupt, incomplete or noisy input data\cite{Wang2023}. In addition to these
naturally occurring impairments, a model can also be disturbed deliberately with
so-called adversarial examples that are a side effect of the internal structures
and training procedures of deep neural networks\cite{Serban2020}. These
carefully calculated patterns are induced into the input data to manipulate the
model's output in the attacker's favor. As these patterns are typically tailored
to specific models and datasets, their transferability, and broader relevance
are often subjects of debate among critics.

This work gives a foundation for answering the question of transferability of
adversarial patterns for real-time object detectors. Specifically, a total
of 28 real-time object detectors are attacked with adversarial patches in an
evasion attack \cite{Chakraborty2021} manner. The object detectors use the pre-trained
weights, optimized on the COCO dataset \cite{Lin2014}. Since most of the more recent
state-of-the-art real-time object detectors are part of the YOLO \cite{Terven2023} family,
RT-DETR \cite{Zhao2024} is the only non-YOLO competitor. The key finding of this
work shows (\autoref{fig:compatibility-matrix}), that the impact of adversarial
patches on the mean average precision is significant high -- even when these
patches are trained on entirely different network architectures.

The fact that adversarial examples are transferable, has already been shown in
different works. Yet, the research focus is mainly on high-frequent pattern
\cite{Wu2018,Naseer2019,Xie2019,Wu2020a,Chow2020,Guo2020,Alvarez2023,Wang2023b},
that usually target image classifiers.
Naseer et al. \cite{Naseer2019} examine cross-domain transferability and
demonstrate the existence of domain-invariant adversaries. The learned
adversarial function on entirely different domains as paintings, cartoons or
medical images can successfully fool classifiers when applied to samples of
ImageNet \cite{Deng2009}.
Cross-dataset transferability is investigated by Staff et al. \cite{Staff2021}.
They attack object detectors that have been learned with disjoint training sets
with the targeted objectness gradient attack \cite{Chow2020}. Their results
suggest, that the more the training data of two models intersect, the more
transferable the attack is.
Alvarez et al. \cite{Alvarez2023} investigate the transferability of different
adversarial attacks between various image classifier architectures. Their
results show, that stronger perturbations and comparable architectures result in
better transferability. Interestingly, they observe that black-box attacks do
not transfer at all.

In addition to directly accessing the transferability of attacks, various works
\cite{Wei2019,Wang2023b,Guo2020,Xie2019,Wu2020a,Wu2018,Zhang2023} have focused
on enhancing the transferability of adversarial examples. For instance, Wang et
al. \cite{Wang2023b} validate empirically, that truncation of the
gradient during back propagation by non-linear layers such as ReLU and
max-pooling making the gradient imprecise regarding the input image to the loss
function. As a result, they propose a novel Backward propagation attack to
increase further the transferability of adversarial examples.
Regarding the transferability of physical attacks against object detectors,
Zhang et al. \cite{Zhang2023} suggest a novel learning method to produce
adversarial patches by redistributing separable attention maps.
A different approach is presented by Wei et al. \cite{Wei2019} based on a
generative adversarial network. Examples generated with the presented
adversarial example generator have a better transferability and can be
generated significantly faster than the compared method.

However, some works \cite{Thys2019, Wu2020, Xu2020, Hoory2020} also conduct
transferability studies for adversarial patches that attack object detectors,
but as the transferability study is only of secondary interest, the discussion
on this topic is consequently quite short.

As this is only a brief review of selected works, the interested reader is
referred to these surveys \cite{Serban2020,Chakraborty2021,Mi2023} for a more
profound overview. However, the presented results reveal a gap in the
literature: the investigation of physical attacks on object detectors. To be
more precise, there is no extensive empirical study on the transferability that
offers more than a vague answer. This work addresses this issue and provides an
empirical study on the transferability of the most popular real-time object
detector family: YOLO \cite{Terven2023}. To the best of our knowledge, this work
is the first that offers an extensive evaluation of SOTA real-time object
detectors of recent years.

The structure of the paper is as follows: \Cref{sec:transferability} gives a
more in-depth analysis of the calculated patches. The experimental setup as well
as a description of the used datasets is given in \cref{sec:setup}. The
results are presented and discussed in \cref{sec:evaluation}. Finally,
\cref{sec:summary} summarizes the main outcomes and gives a brief outlook.

To ensure the reproducibility of our results, we provide the code used to create
the patches and the evaluation code in our
repository\footnote{\url{https:\\\\github.com/JensBayer/transferability}}. 

\section{Transferability of Adversarial Patches}
\label{sec:transferability}
Instead of directly going into the in-depth analysis of the patch properties,
the optimization process of the patches is presented first. Afterward, the
patches are analyzed with their corresponding t-SNE embeddings and compared
statistically.

\subsection{Patch Optimization}
The patch optimization follows the procedure presented by Thys et al.
\cite{Thys2019}. Each investigated model uses the official provided pretrained
weights on the COCO dataset \cite{Lin2014}. The resolution of the patches is
$256 \times 256$ pixel. Each patch is initialized with random noise in the range
[0,1] and is optimized with the AdamW \cite{Loshchilov2019} optimizer over 100
epochs with an initial learning rate of $0.01$ and a learn rate reduction by
factor 10 each 25th epoch. During training, the patches are augmented by a
random resize in the range of $[0.75, 1.0]$ of the original bounding box size, a
random color jitter, random perspective and a random rotation in the range of
$\pm 30$ degree. To smoothen the patch and keep values in a valid range, a
smoothening term $L_s$ and a validity loss $L_v$ are calculated. The final loss
\begin{equation}
    Loss = \lambda_s L_s + \lambda_v L_v + \lambda_m L_m
\end{equation} is then given by the weighted sum of the smoothening, validity
and target loss. The investigated patches are used in an evasion
attack setting. Therefore, the optimization objective is to minimize the network
activations when the trained patch is applied on objects of interest. As
different architectures are used to generate patches, the target loss $L_m$
differs slightly for some model $m$. 

\subsubsection{YOLOv7}
YOLOv7\cite{Wang2022} is the oldest investigated architecture group and still uses an
objectness score to provide information about whether a certain anchor box
contains an object at all. The loss term 
\begin{equation}
    L_{v7} = \max(obj\, scores)
\end{equation}
is therefore set to be the maximum objectness score over all anchor boxes. Furthermore,
the maximum is extracted before the final sigmoid activation and non-maximum suppression
are applied.

\subsubsection{Yolov8/v9/NAS/RT-DETR}
For most of the investigated models (YOLOv8\cite{Ultralytics2023},
YOLOv9\cite{Wang2024a}, YOLO-NAS\cite{supergradients2021},
RT-DETR\cite{Zhao2024}) the loss term
\begin{equation}
    L_{v8} = L_{v9} = L_{NAS} = \max(det\,scores)
\end{equation}
is given by the maximum class prediction score over all detected bounding
boxes. Similar to the YOLOv7 group, the extraction of the maximum happens before
the final sigmoid activation and non-maximum suppression are applied.

\subsubsection{YOLOv10}
In comparison to the previous architecture groups, YOLOv10\cite{Wang2024} does
not require a non-maximum suppression. Instead, a dual label assignment is
used. As a result, the loss term
\begin{equation}
    L_{v10} = \max(\text{one2many\,det\,scores}) + \max(\text{one2one\,det\,scores})
\end{equation}
 is calculated in the same manner as the previous methods, but considers both heads.

\section{Patch Properties}
\subsection{Inception t-SNE Analysis}
Instead of directly embedding the patches pixelwise with a t-distributed
stochastic neighbor embedding (t-SNE), a more sophisticated analysis variant is
chosen. Inspired by the Inception Score \cite{Salismans2016} and motivated by
the fact, that a pixelwise comparison of patches fails to capture similar
structures located differently within patches, the patches are first activated
with the Inceptionv3\cite{Szegedy2016} network. The features of the last
convolutional layer of the network are used as the input for the t-SNE
analysis.

\begin{figure}
    \centering
    \includegraphics[width=0.8\linewidth]{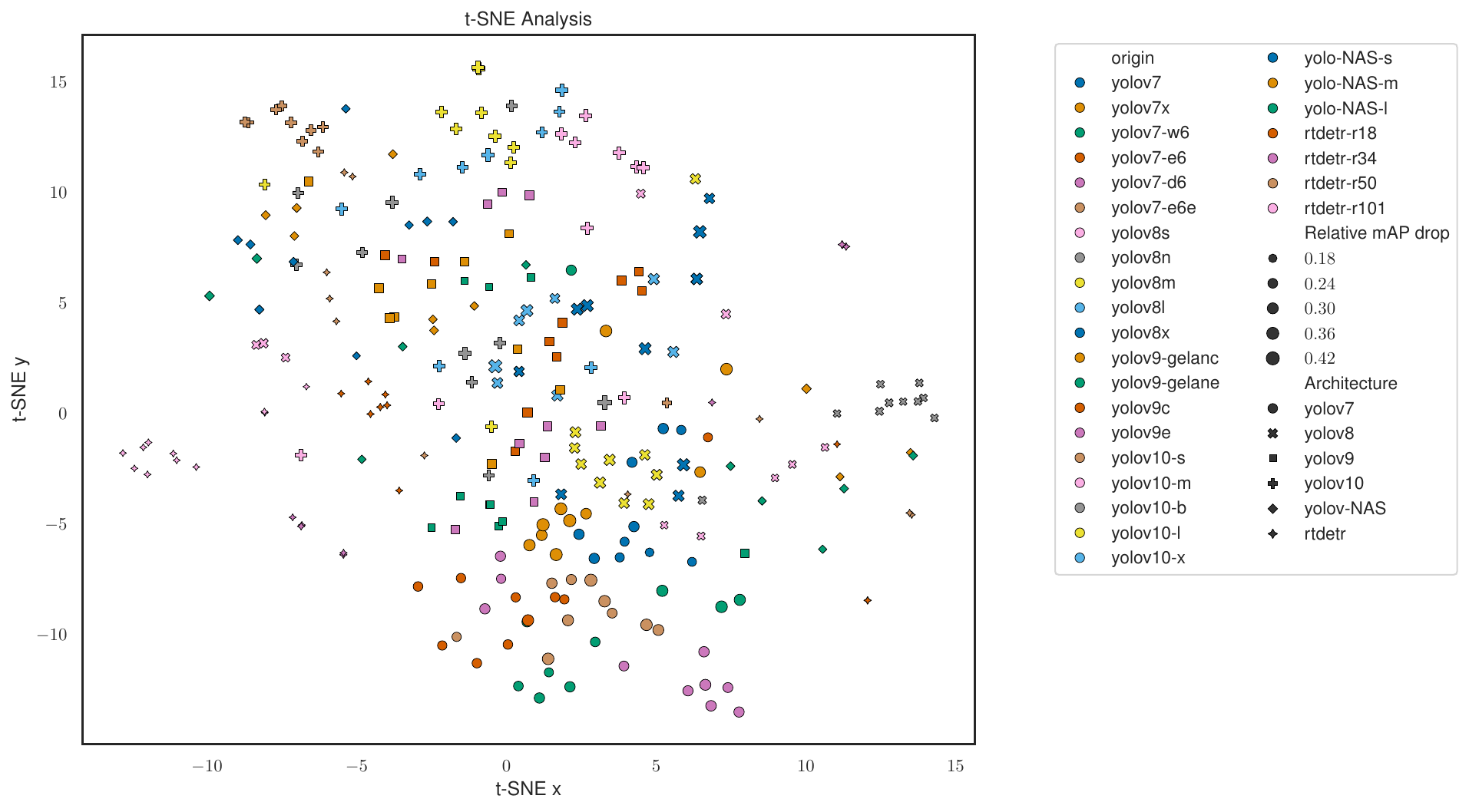}
    \caption{Embedded Inceptionv3 features of the patches via t-SNE. The color
    of each data point corresponds to the origin, the symbol to the general
    architecture group, and the size of the symbol to the relative mAP drop. The
    mAP drop refers to the INRIA Person test set.}
    \label{fig:inception_tsne}
\end{figure}

Each data point of the resulting plot (see \autoref{fig:inception_tsne}) refers
to a single patch. The color determines the origin, the symbol provides
information about the general architecture used to train, and the size of the
symbol encodes the relative mean average precision (mAP) \cite{Lin2014} drop of
the patch. By using this representation, the clusters of the different
architecture groups and networks become visible. For example, almost all YOLOv7
patches are located below $y=-2$ and $-5 < x < 10$. In addition to the obvious
clusters, the relative mAP drop of different clusters becomes visible. The
larger the symbol, the higher the mAP drop is. Clusters that are further away
from the center result in a lower mAP drop and perform therefore worse, compared
to the more central ones. These more distant patches originate from the RT-DETR
architectures and the YOLOv8-s network. As depicted in
\autoref{fig:compatibility-matrix}, the patches trained with these networks
suffer from a low overall mAP drop.

\subsection{Histogram Analysis}
\begin{figure}[tb]
    \centering
    \includegraphics[width=0.18\linewidth]{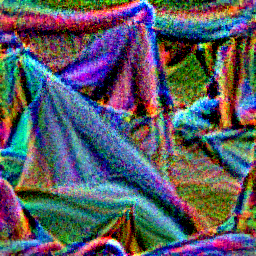}
    \includegraphics[width=0.18\linewidth]{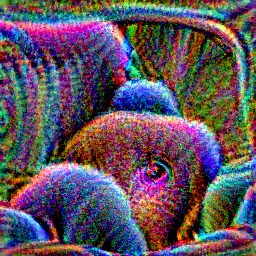}
    \includegraphics[width=0.18\linewidth]{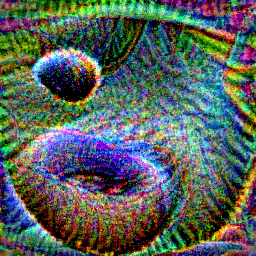}
    \includegraphics[width=0.18\linewidth]{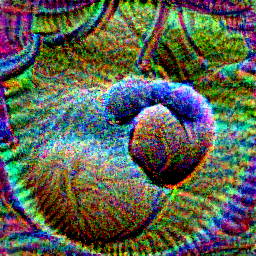}
    \includegraphics[width=0.18\linewidth]{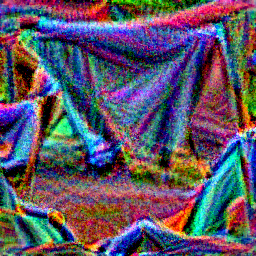}
    \includegraphics[width=0.18\linewidth]{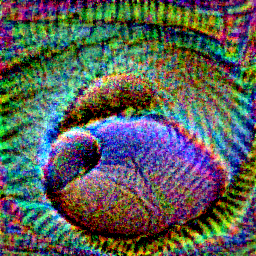}
    \includegraphics[width=0.18\linewidth]{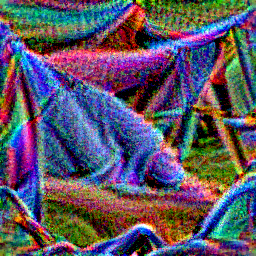}
    \includegraphics[width=0.18\linewidth]{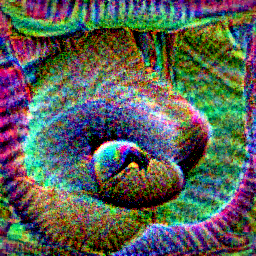}
    \includegraphics[width=0.18\linewidth]{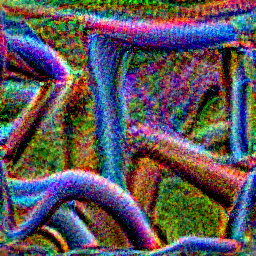}
    \includegraphics[width=0.18\linewidth]{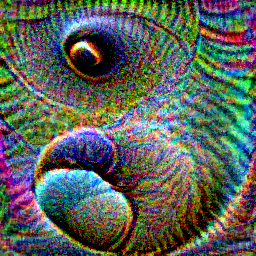}
    \caption{Patches that have been optimized with the YOLOv8-s network. Certain
    similarities in terms of color and shape are clearly visible.}
    \label{fig:v8s-patches}
\end{figure}

\begin{figure}[tb]
    \centering
    \includegraphics[width=\linewidth]{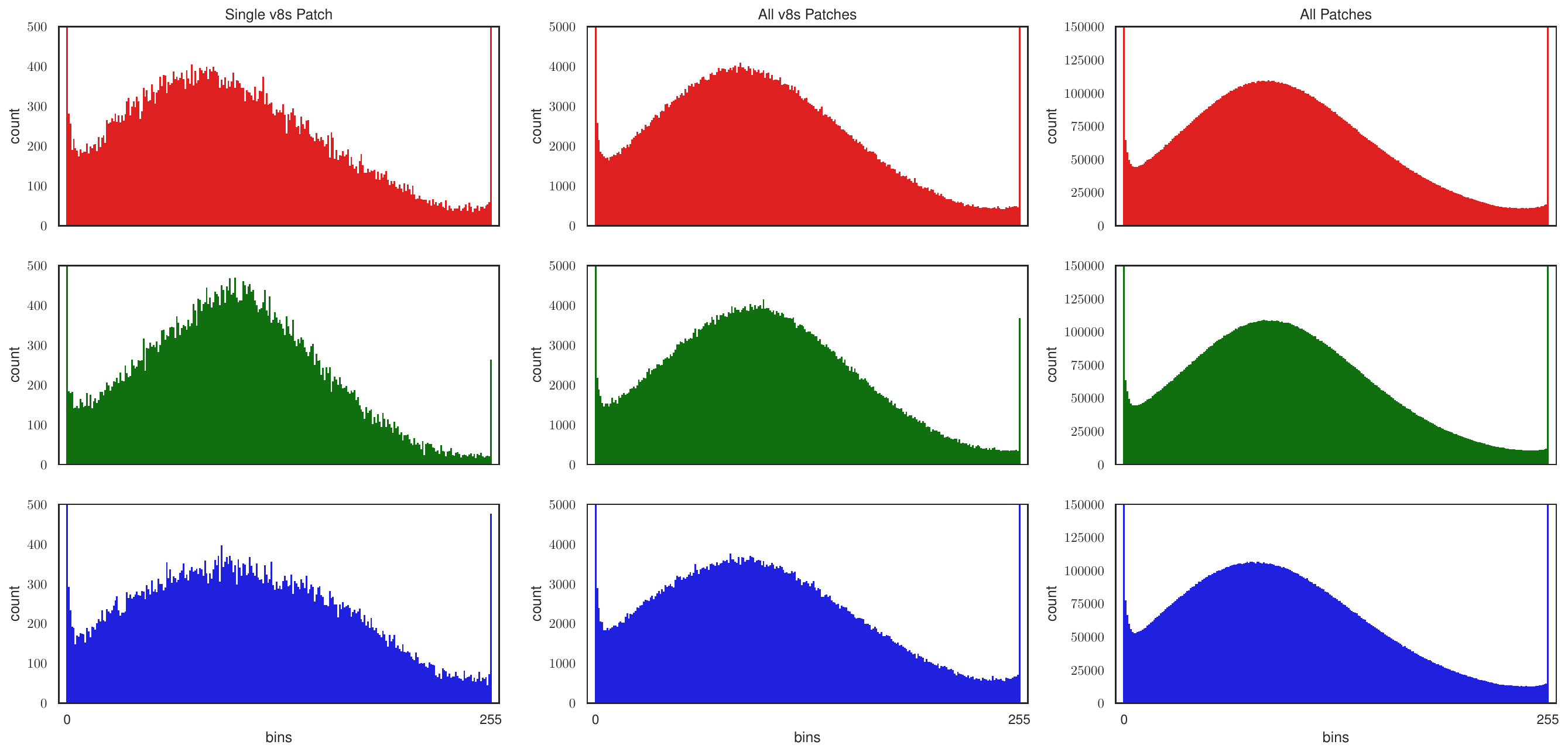}
    \caption{Histograms of the RGB channels. The first column shows the
    histograms of a single patch. The second column, the histograms of all
    patches optimized with the YOLOv8-s network. The third, the histograms of
    all optimized patches, regardless of the network used to optimize a patch.}
    \label{fig:rgb_analysis}
\end{figure}

\begin{figure}
    \centering
    \includegraphics[width=\linewidth]{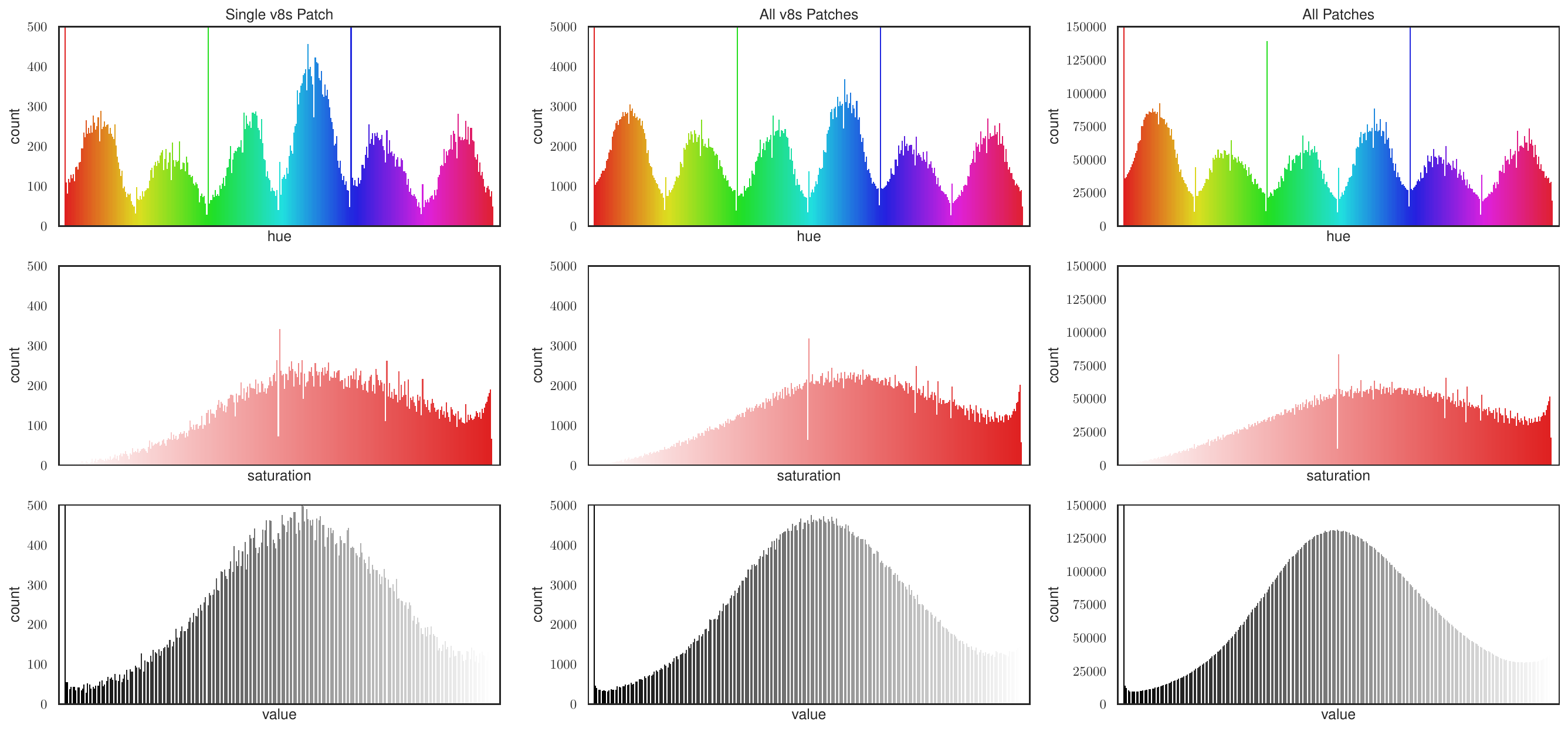}
    \caption{Histograms of the HSV-channels. As in \autoref{fig:rgb_analysis},
    the first column shows the histograms of a single patch, the second the
    histograms of all patches optimized with the YOLOv8-s network and the third
    the histograms of all optimized patches.}
    \label{fig:hsv_analysis}
\end{figure}

To check if the YOLOv8-s patches differ in a certain way regarding the overall
color distribution, the histograms of both, the RGB (see
\autoref{fig:rgb_analysis}) and the HSV (see \autoref{fig:hsv_analysis}) values
are analyzed. Both plots contain three columns and three rows. Each row depicts
the histogram of a certain RGB or HSV channel. The first column refers
to the first patch (\autoref{fig:v8s-patches}, top left) generated with the YOLOv8-s
network. The second column to the histogram of all ten patches
(\autoref{fig:v8s-patches}), generated with the YOLOv8-s network. The third row
is the histogram of all 280 patches regarding the architecture.

All three RGB channels show spikes at the lowest (0) and highest (255) values. This is
true whether it is the single YOLOv8-s patch, all YOLOv8-s patches or all generated
patches. The distributions of the histograms regarding the channels are also
quite similar, as the values in \autoref{tab:rgb_stats} show. 
However, aside from these spikes, the histograms offer limited insight as they
do not indicate which colors dominate in a patch. To gain a clearer
understanding of color distribution, an additional HSV analysis is performed, as
illustrated in \autoref{fig:hsv_analysis}. This way, the color distributions
among the patches becomes visible. Similar to the previous spikes, the hue
histograms in the first row show three spikes for red, green, and blue. Again,
the HSV histograms of the single YOLOv8-s patch are comparable with the
histograms of all YOLOv8-s patches and the histograms of all patches. Yet,
differences are visible, especially in the hue and saturation histograms.

\begin{table}[tb]
    \centering
    \begin{tabular}{l l r c c c c}
        \hline
         Channel & Source & Mean $\pm$ Std. & Median & Skewness & Kurtosis \\
         \hline
         \multirow{3}*{Red} & Single v8n & $99\pm57$ & $93$ & $0.44$ & $-0.42$ \\
         & All v8n & $100\pm56$ & $95$ & $0.42$ & $-0.35$ \\
         & All & $97\pm57$ & $91$ & $0.46$ & $-0.29$ \\

         \multirow{3}*{Green} & Single v8n & $102\pm52$ & $100$ & $0.26$ & $-0.33$ \\
         & All v8n & $103\pm55$ & $100$ & $0.33$ & $-0.41$ \\
         & All & $99\pm55$ & $95$ & $0.44$ & $-0.28$ \\
         
         \multirow{3}*{Blue} & Single v8n & $109\pm60$ & $105$ & $0.25$ & $-0.72$ \\
         & All v8n & $103\pm59$ & $97$ & $0.39$ & $-0.50$ \\
         & All & $97\pm57$ & $91$ & $0.51$ & $-0.30$ \\
         \hline
    \end{tabular}
    \caption{Statistics of the RGB histograms. Only values in the range [1,254]
    are considered. For each histogram, the mean, standard deviation, median,
    skewness, and kurtosis are calculated.}
    \label{tab:rgb_stats}
\end{table}

\section{Experimental Setup}
\label{sec:setup}
To ensure a fair comparison, all patches are embedded at the same positions and
the same scaling within the bounding boxes of objects of interest. That is, the
center of a ground truth bounding box with a scale equal to 0.75 times the
shorter side of the bounding box. To have a reference on how important the
covered area is, the performance of the networks is measured, when the same area
is covered with different grayscale levels. Similar to the optimization
procedure, the evaluated networks use the official provided pretrained weights
on the COCO dataset. 
The relative mean average precision drop is used to quantify the performance of
a patch. First, the mAP of a detector is calculated when no patches are present
in the test data. Then, the mAP is calculated again, when the bounding boxes of
objects of interests contain the patches. Finally, the difference between the
two mAP values is calculated.

The patches are optimized for the exemplary reference category \enquote{Person}.
As a result, the training dataset is the INRIA Person Dataset \cite{Dalal2005}. The
evaluation is performed on two datasets. In addition to the test set of the
INRIA Person dataset, images of the COCO test set that contain objects of the
target category are evaluated as well.

\section{Evaluation}
\label{sec:evaluation}
\begin{figure}[!htb]
    \centering
    \includegraphics[width=0.6\linewidth]{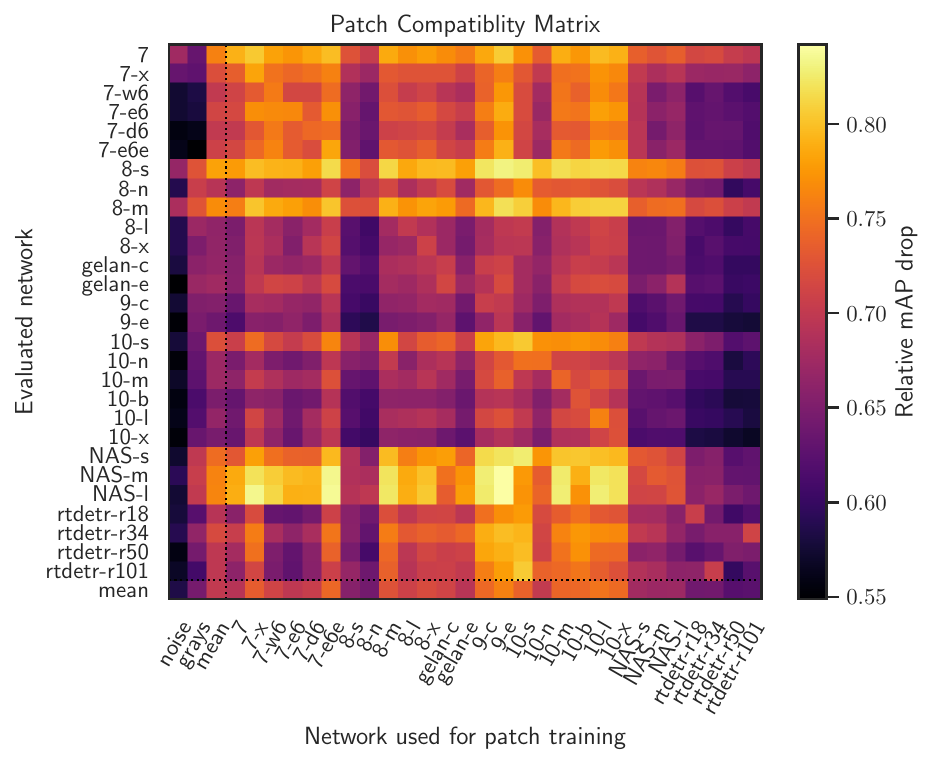}
    \caption{Compatibility matrix of the networks when evaluated on the COCO
    test dataset. Each pixel represents the mean average precision (mAP) drop of
    a set of patches on the test set for a specific network. The brighter the
    color, the higher the relative mAP drop of the patch. A bright vertical and
    a dark horizontal line for a single network is desirable.}
    \label{fig:coco-compatibility}
\end{figure}

\autoref{fig:compatibility-matrix} and \autoref{fig:coco-compatibility}
summarize the evaluation results. Networks that are used to train a patch are
mapped on the x-axis. The evaluated networks are mapped on the y-axis. 
In addition to the optimized patches, we also checked the impact of noise or
grayscale patches on the evaluated networks. The results are located in the
first two columns, next to the rowwise mean. The color of a single entry of the
heatmap encodes the relative mAP drop a patch causes. The brighter the color,
the higher the mAP drop is.

When inspecting the heatmap in \autoref{fig:compatibility-matrix}, the rightmost
dark vertical block is prominent. Here, the overall tone is remarkably darker
than the rest of the heatmap. Only the YOLOv8-s and YOLOv8-n networks perform
similarly. This block belongs to the YOLO-NAS and RT-DETER architecture groups.
The patches trained with these architectures are by far less transferable in
comparison to the YOLOv9 or YOLOv10 architectures.
Furthermore, almost every architecture group optimizes more efficient patches
in terms of attacking the YOLO-NAS architecture, than patches trained with the
architecture per se. The exception here is again the RT-DETR and the YOLOv8-s
and YOLOv8-n architecture. The same is true for the RT-DETR architecture group.
Yet, this group is more resistant against patches optimized with YOLOv7 and
YOLOv8.

The highest mAP drops are achieved with the YOLOv10 and YOLOv9 architectures.
Moreover, networks of these architecture groups are more resistant against
patches that are optimized with networks of the remaining architecture groups,
as long as the networks are large enough. A general observation is, that a larger
network usually produces better patches than a smaller network of the same
architecture group. This is also sometimes, but not always across all
architectures, the case (e.g., YOLOv7-e6e).

The results achieved on the COCO test set (see \autoref{fig:coco-compatibility})
are consistent with those from the INRIA Person dataset. Patches produced with
YOLO-NAS, RT-DETR and the two small YOLOv8 networks perform much worse than the
rest. Surprisingly, the YOLOv7 networks are much more sensitive to the patches
compared to the results achieved with the INRIA Person test set. In general, the
heatmap is much brighter. Moreover, the scale covers a smaller range but starts
at a higher value. This means that the COCO test set is more complex and that
the impact on the detection performance caused by the patches, transfers, even
when the patches are trained on an entirely different dataset.

Regarding the performance of the networks when uniform noise patches are
present, the impact on the INRIA Person dataset is negligible, while the mAP drop
for the COCO dataset is much higher. When attacked with different shades of
gray, the impact on the INRIA Person dataset is by far less low than on the
COCO dataset. The cause for the investigated behavior is probably the higher
complexity of the COCO dataset. Nonetheless, an interesting finding is that all
networks seem to be sensible to grayscale patches with the value of 0.5. We
hypothesize that this sensitivity arises from an unintended side effect of the
training procedures, where such a grayscale value is usually used for padding or
to fill gaps between images when using mosaic training.

\section{Summary}
\label{sec:summary}
This work analyzes the transferability of adversarial patches between multiple
real-time object detector networks by comparing the relative mean average
precision. A total of 280 patches have been optimized and evaluated with 28
different models. INRIA Person is used as the training dataset for the patches,
while the evaluation is performed on the test sets of INRIA Person and COCO. The
results indicate that patches that are optimized with larger models have a
higher transferability between different models than patches that are optimized
with smaller models. The YOLO-NAS architecture provides robustness against these
patches, while the best robustness is provided by the more recent YOLOv9 and
YOLOv10 architectures. Similarly, the best transferability is achieved
with the larger YOLOv9 and YOLOv10 architectures.
Future research should explore the hypothesis that the networks' sensitivity to
specific grayscale patches may be an unintended consequence of the training
procedure.

\section*{ACKNOWLEDGMENTS}       
 
This work was developed in Fraunhofer Cluster of Excellence \enquote{Cognitive Internet Technologies}.
\bibliography{references} 

\begin{thebibliography}{10}

\bibitem{Wang2023}
S.~Wang, R.~Veldhuis, and N.~Strisciuglio, ``{The Robustness of Computer Vision
  Models against Common Corruptions: a Survey},'' {\em arXiv Prepr.} ,
  pp.~1--23, may 2023.

\bibitem{Serban2020}
A.~Serban, E.~Poll, and J.~Visser, ``{Adversarial Examples on Object
  Recognition: A Comprehensive Survey},'' {\em ACM Computing Surveys}~{\bf
  53}(3), 2020.

\bibitem{Chakraborty2021}
A.~Chakraborty, M.~Alam, V.~Dey, A.~Chattopadhyay, and D.~Mukhopadhyay, ``{A
  survey on adversarial attacks and defences},'' {\em CAAI Trans. Intell.
  Technol.}~{\bf 6}(1), pp.~25--45, 2021.

\bibitem{Lin2014}
T.-Y. Lin, M.~Maire, S.~Belongie, J.~Hays, P.~Perona, D.~Ramanan,
  P.~Doll{\'a}r, and C.~L. Zitnick, ``Microsoft coco: Common objects in
  context,'' pp.~740--755, 2014.

\bibitem{Terven2023}
J.~Terven and D.~Cordova-Esparza, ``{A Comprehensive Review of YOLO: From
  YOLOv1 to YOLOv8 and Beyond},'' in {\em arXiv preprint},  pp.~1--27, 2023.

\bibitem{Zhao2024}
Y.~Zhao, W.~Lv, S.~Xu, J.~Wei, G.~Wang, Q.~Dang, Y.~Liu, and J.~Chen, ``Detrs
  beat yolos on real-time object detection,'' in {\em CVPR},  pp.~16965--16974,
  June 2024.

\bibitem{Wu2018}
L.~Wu, Z.~Zhu, C.~Tai, and W.~E, ``{Understanding and Enhancing the
  Transferability of Adversarial Examples},'' in {\em arXiv preprint},  (2015),
  pp.~1--15, 2018.

\bibitem{Naseer2019}
M.~Naseer, S.~Khan, M.~H. Khan, F.~S. Khan, and F.~Porikli, ``{Cross-domain
  transferability of adversarial perturbations},'' {\em Adv. Neural Inf.
  Process. Syst.}~{\bf 32}(NeurIPS), pp.~1--11, 2019.

\bibitem{Xie2019}
C.~Xie, Z.~Zhang, Y.~Zhou, S.~Bai, J.~Wang, Z.~Ren, and A.~L. Yuille,
  ``{Improving transferability of adversarial examples with input diversity},''
  {\em Proc. IEEE Comput. Soc. Conf. Comput. Vis. Pattern Recognit.}~{\bf
  2019-June}, pp.~2725--2734, 2019.

\bibitem{Wu2020a}
W.~Wu, Y.~Su, X.~Chen, S.~Zhao, I.~King, M.~R. Lyu, and Y.-W. Tai, ``Boosting
  the transferability of adversarial samples via attention,'' in {\em
  Proceedings of the IEEE/CVF Conference on Computer Vision and Pattern
  Recognition (CVPR)},  June 2020.

\bibitem{Chow2020}
K.-h. Chow, L.~Liu, M.~E. Gursoy, S.~Truex, W.~Wei, and Y.~Wu, ``{TOG :
  Targeted Adversarial Objectness Gradient Attacks on Real-time Object
  Detection Systems},'' in {\em arXiv preprint},  2020.

\bibitem{Guo2020}
Y.~Guo, Q.~Li, and H.~Chen, ``Backpropagating linearly improves transferability
  of adversarial examples,'' {\em Advances in neural information processing
  systems}~{\bf 33}, pp.~85--95, 2020.

\bibitem{Alvarez2023}
E.~Alvarez, R.~Alvarez, and M.~Cazorla, ``{Exploring Transferability on
  Adversarial Attacks},'' {\em IEEE Access}~{\bf 11}(October),
  pp.~105545--105556, 2023.

\bibitem{Wang2023b}
X.~Wang, K.~Tong, and K.~He, ``Rethinking the backward propagation for
  adversarial transferability,'' in {\em Proceedings of the 37th International
  Conference on Neural Information Processing Systems},  {\em NIPS '23}, Curran
  Associates Inc., (Red Hook, NY, USA), 2024.

\bibitem{Deng2009}
J.~Deng, W.~Dong, R.~Socher, L.-J. Li, K.~Li, and L.~Fei-Fei, ``Imagenet: A
  large-scale hierarchical image database,'' in {\em 2009 IEEE conference on
  computer vision and pattern recognition},  pp.~248--255, Ieee, 2009.

\bibitem{Staff2021}
A.~M. Staff, J.~Zhang, J.~Li, J.~Xie, E.~A. Traiger, J.~A. Glomsrud, and K.~B.
  Karolius, ``{An Empirical Study on Cross-Data Transferability of Adversarial
  Attacks on Object Detectors},'' {\em CEUR Workshop Proceedings}~{\bf 3125},
  pp.~38--52, 2021.

\bibitem{Wei2019}
X.~Wei, S.~Liang, N.~Chen, and X.~Cao, ``{Transferable adversarial attacks for
  image and video object detection},'' {\em IJCAI Int. Jt. Conf. Artif.
  Intell.}~{\bf 2019-Augus}, pp.~954--960, 2019.

\bibitem{Zhang2023}
Y.~Zhang, Z.~Gong, Y.~Zhang, K.~Bin, Y.~Li, and J.~Qi, ``{Boosting
  transferability of physical attack against detectors by redistributing
  separable attention},'' {\em Pattern Recognit.}~{\bf 138}, p.~109435, 2023.

\bibitem{Thys2019}
S.~Thys, W.~V. Ranst, and T.~Goedeme, ``{Fooling automated surveillance
  cameras: Adversarial patches to attack person detection},'' {\em CVPR
  Work.}~{\bf 2019-June}, pp.~49--55, 2019.

\bibitem{Wu2020}
Z.~Wu, S.~N. Lim, L.~S. Davis, and T.~Goldstein, ``{Making an Invisibility
  Cloak: Real World Adversarial Attacks on Object Detectors},'' {\em Lect.
  Notes Comput. Sci.}~{\bf 12349 LNCS}, pp.~1--17, 2020.

\bibitem{Xu2020}
K.~Xu, G.~Zhang, S.~Liu, Q.~Fan, M.~Sun, H.~Chen, P.~Y. Chen, Y.~Wang, and
  X.~Lin, ``{Adversarial T-Shirt! Evading Person Detectors in a Physical
  World},'' {\em Lect. Notes Comput. Sci. (including Subser. Lect. Notes Artif.
  Intell. Lect. Notes Bioinformatics)}~{\bf 12350 LNCS}, pp.~665--681, 2020.

\bibitem{Hoory2020}
S.~Hoory, T.~Shapira, A.~Shabtai, and Y.~Elovici, ``Dynamic adversarial patch
  for evading object detection models,'' {\em arXiv preprint arXiv:2010.13070}
  , 2020.

\bibitem{Mi2023}
J.-x. Mi, X.-d. Wang, L.-f. Zhou, and K.~Cheng, ``{Adversarial examples based
  on object detection tasks: A survey},'' {\em Neurocomputing}~{\bf 519},
  pp.~114--126, jan 2023.

\bibitem{Loshchilov2019}
I.~Loshchilov and F.~Hutter, ``{Decoupled weight decay regularization},'' {\em
  7th Int. Conf. Learn. Represent. ICLR 2019} , 2019.

\bibitem{Wang2022}
C.-Y. Wang, A.~Bochkovskiy, and H.-Y.~M. Liao, ``{YOLOv7: Trainable
  Bag-of-Freebies Sets New State-of-the-Art for Real-Time Object Detectors},''
  in {\em 2023 IEEE/CVF Conf. Comput. Vis. Pattern Recognit.},  pp.~7464--7475,
  IEEE, jun 2023.

\bibitem{Ultralytics2023}
G.~Jocher, A.~Chaurasia, and J.~Qiu, ``{Ultralytics YOLO},'' Jan. 2023.

\bibitem{Wang2024a}
C.-Y. Wang, I.-H. Yeh, and H.-Y.~M. Liao, ``Yolov9: Learning what you want to
  learn using programmable gradient information,'' {\em arXiv preprint
  arXiv:2402.13616} , 2024.

\bibitem{supergradients2021}
S.~Aharon, {Louis-Dupont}, {Ofri Masad}, K.~Yurkova, {Lotem Fridman}, {Lkdci},
  E.~Khvedchenya, R.~Rubin, N.~Bagrov, B.~Tymchenko, T.~Keren, A.~Zhilko, and
  {Eran-Deci}, ``Super-gradients,'' 2021.

\bibitem{Wang2024}
A.~Wang, H.~Chen, L.~Liu, K.~Chen, Z.~Lin, J.~Han, and G.~Ding, ``{YOLOv10:
  Real-Time End-to-End Object Detection},'' in {\em arXiv preprint},
  pp.~1--18, 2024.

\bibitem{Salismans2016}
T.~Salimans, I.~Goodfellow, W.~Zaremba, V.~Cheung, A.~Radford, and X.~Chen,
  ``Improved techniques for training gans,'' in {\em NIPS},  {\em NIPS'16},
  p.~2234–2242, Curran Associates Inc., (Red Hook, NY, USA), 2016.

\bibitem{Szegedy2016}
C.~Szegedy, V.~Vanhoucke, S.~Ioffe, J.~Shlens, and Z.~Wojna, ``{Rethinking the
  Inception Architecture for Computer Vision},'' {\em Proc. IEEE Comput. Soc.
  Conf. Comput. Vis. Pattern Recognit.}~{\bf 2016-December}, pp.~2818--2826,
  2016.

\bibitem{Dalal2005}
N.~Dalal and B.~Triggs, ``{Histograms of Oriented Gradients for Human
  Detection},'' in {\em CVPR},   {\bf 1}, pp.~886--893, IEEE, 2005.

\end{thebibliography}
\bibliographystyle{spiebib} 

\end{document}